\documentclass[11pt,a4paper]{article}
\usepackage[hyperref]{emnlp-ijcnlp-2019}
\usepackage{latexsym}
\usepackage{times}
\usepackage{url}

\usepackage{xspace}
\usepackage{booktabs}
\usepackage{color}
\usepackage{graphicx}
\usepackage{tabulary}
\usepackage{enumitem}
\usepackage{url}
\usepackage{amsfonts}

\newenvironment{tightitemize}%
  {\begin{itemize}[topsep=0pt, partopsep=0pt] %
    \setlength{\itemsep}{0pt}%
    \setlength{\parskip}{0pt}%
    }%
  {\end{itemize}}
  \usepackage{booktabs}
\usepackage{subcaption}
\usepackage{lipsum}

  \usepackage{color}
  {\begin{enumerate}â \footnotesize%
    \setlength{\itemsep}{0pt}%
    \setlength{\parskip}{0pt}%
    }%
  {\end{enumerate}}


\usepackage{microtype}
\usepackage{multirow}
\usepackage{verbatim}
\usepackage{amsmath,amsthm,amssymb}
\usepackage{array}
\usepackage[scaled=0.86]{helvet}
\usepackage{ifthen}
\usepackage{courier}
\usepackage[boxed]{algorithm}
\usepackage{caption}
\usepackage{algpseudocode}
\aclfinalcopy

\belowdisplayskip \abovedisplayskip

\usepackage{amsmath}

\usepackage{CJKutf8}

\title{Query-Based Named Entity Recognition}

\author{Yuxian Meng, Xiaoya Li, Zijun Sun and Jiwei Li \\
Shannon.AI\\
   {\{yuxian\_meng, xiaoya\_li, zijun\_sun,  jiwei\_li\}@shannonai.com}
}

\date{}

\begin{document}
\begin{CJK*}{UTF8}{gbsn}
\maketitle
\begin{abstract}
In this paper, we propose a new strategy for the task of named entity recognition (NER). We cast
 the task as a query-based machine reading comprehension task: e.g., the task of extracting entities with \textsc{per} is formalized as answering the question of ``{\it which person is mentioned in the text ?}". 
Such a strategy comes with the  advantage that it solves the long-standing issue of handling overlapping or nested entities
  (the same token that participates in more than one entity categories) 
   with sequence-labeling techniques for NER.
   Additionally, since the query encodes informative prior knowledge, this strategy facilitates the process of entity extraction, leading to better performances.

We experiment the proposed model on five widely used NER datasets on English and Chinese, including MSRA, Resume, OntoNotes, ACE04 and ACE05. The proposed model sets new SOTA results on all of these datasets. \footnote{Please refer to the full version of this paper: A Unified MRC Framework for Named Entity Recognition  \url{https://arxiv.org/pdf/1910.11476.pdf}}

\end{abstract}

\section{Introduction}
Named entity recognition （NER） is a basic  task in building natural language processing (NLP) systems. 
The task is traditionally formalized as a sequence labeling problem, in which an algorithm needs to assign a tagging class to each word or
character  within a sequence. 
Depending on whether the prediction of the label is made based on its proceeding/surrounding labels, existing models can be divided into two major categories: (1) autoregressive ones such as CRF-based models  \cite{lample2016neural, ma2016end, chiu2016ner, latticelstm2018} and (2) non-autoregressive ones such as BERT  
\cite{devlin2018bert}. 

Existing approaches, both autoregressive and non-autoregressive ones, come with some intrinsic drawbacks at both the formalization level and the algorithmic level. 
At the formalization level, most current models are incapable of handling overlapping or nested entities  \cite{kim2003genia, finkel2009nested, luroth2015, kati2018nested}. 
This is because 
of the fact that one token can only be assigned to one tag category. 
At the algorithmic level, tagging classes are merely  indexes and  do not
encode any prior information about entity categories. This lack of clarity
on what to extract leads to inferior performances.

 In this paper, we propose a new 
paradigm to extract named entities. 
We formalize the task as a question answering task in machine reading comprehension: each entity type is characterized by a natural language query, and entities are extracted by answering these queries given context. 
 For example, the task of assigning the PER label to  {\em Washington\/}  in {\em [Washington] was born into slavery on the farm of James Burroughs\/} 
 is formalized as answering the question {\it which person is mentioned in the text ?}.
 
 Such a type of formalization to a large extent solves the aforementioned issues: (1) the model is able to naturally handle the entity 
overlapping issue:
regarding different entity categories, the model extracts corresponding entity spans by answering different (and independent) questions; (2)  
the  query
encodes significant prior information about the entity class to extract. For example, the semantic relatedness between the query 
 {\em who is mentioned in the text} and the PER entities facilitatew the extracting process, potentially leading to better performances; (3) We are able to take advantages of current well-developed sophisticated MRC models.
  
  Using the proposed strategy, we are able to achieve SOTA results on five NER datasets in English and Chinese, including MSRA, RESUME, Chinese OntoNotes, ACE04 and ACE05. 

\section{Related Work}
\subsection{Named Entity Recognition} 
Traditional  sequence  labeling  models
use CRFs \cite{lafferty2001conditional,sutton2007dynamic} as a backbone.
The first work using neural models for NER goes back to 2003, when
\newcite{hammerton2003named} attempted to solve the problem using unidirectional LSTMs. 
\newcite{collobert2011natural} presented the  CNN-CRF structure, augmented   with character embeddings by \newcite{santos2015boosting}. 
\newcite{lample2016neural}  explored neural structures for NER, in which
the bidirectional LSTMs are combined with CRFs  
 with features based on 
 character-based word
representations  and unsupervised word representations.
\newcite{ma2016end} 
 and \newcite{chiu2016named}  
 used a character CNN to extract features from characters.
  Recent large-scale language model pertaining methods such as BERT \cite{devlin2018bert} and Elmo \cite{elmo2018} further enhance the performance of NER, yielding  state-of-the-art performances.

Nested NER refers to a situation in  which overlapping or nested entity mentions exist.
This phenomenon was first noticed by \newcite{kim2003genia}, in which rules were used to identify overlapping mentions. 
\newcite{finkel2009nested} made the assumption that one mention is fully contained by the other when they overlap 
and built a model to extract nested entity mentions based on parse trees. 
\newcite{luroth2015} proposed to use mention hypergraphs for recognizing overlapping mentions. 
\newcite{xu2017fofe} utilize a local classifier that runs on every possible span to detect overlapping mentions and 
\newcite{kati2018nested}
used neural model to learn the  hypergraph representations for nested entities. 
\subsection{Machine Reading Comprehension}
MRC models extract answer spans from passages given questions \cite{seo2016bidirectional,wang2016multi,wang2016machine,xiong2016dynamic,xiong2017dcn,wang2016multi,shen2017reasonet,chen2017reading}. 
The task can be formalized as two multi-class classification tasks, i.e., predicting the starting and ending positions of the answer spans given questions. 

Many NLP tasks can be transformed to the task of question answering. For example, \citet{levy2017zero} 
transformed the task of relation extraction to a QA task:
each relation type R(x,y) can be parameterized as a question q(x) whose answer is $y$. For example, the relation \textsc{educated-at}
can be mapped to  “Where did x study?”. 
Given a question $q(x)$,
if a non-null answer $y$ can be extracted from a sentence, it means 
the relation label for the current sentence is R. 
\newcite{mccann2018natural} transforms 
 NLP tasks such as summarization or sentiment analysis 
    into question answering, 
For example, the task of summarization can  be formalized as answering the question {\it What is the
summary?}.

    \section{Model}
\subsection{System Overview}
Given a word or character sequence $X = \{x_1, x_2, ..., x_n\}$, where $n$ denotes the length of the sequence, we need to assign each token $x_i$ a label $y_i \in Y$, indicating the label for $x_i$. $y_i$ is selected from the predefined list $Y$ for tag types (e.g., PER, LOC, etc).

For each tag type $y\in Y$, it is associated with a natural language question $q_y$. 
Given  $X$ and question $q_y$,
the MRC model is run 
to predict the starting index $start\in [1,n]$ and  the ending index $end\in [1,n]$.  
This leads 
to the extracted answer span [$x_{start}, x_{start+1}, ..., x_{end-1}, x_{end}]$. 
The MRC model allows returning a special NULL token, indicating that no substring within $x$ should be used as the answer to $q_y$. 
If the return is not NULL, the label for  [$x_{start}, x_{start+1}, ..., x_{end-1}, x_{end}]$ will be changed to $y$. We iterate this process for all tagging categories until the end. The overview of the algorithm is shown in Algorithm \ref{alg}.

\algrenewcommand{\algorithmicrequire}{\textbf{Input:}}
\algrenewcommand{\algorithmicensure}{\textbf{Output:}}
\newcommand{\To}{{\bf to }}
\newcommand{\IF}{{\bf if }}
\newcommand{\DO}{{\bf do }}
\newcommand{\ENDIF}{{\bf endif}}
\begin{algorithm}[t]
\small
\begin{algorithmic}[1]
\Require sequence $X=\{x_1, x_2,...,x_n\}$, 
QuestionTemplates, 
\Ensure sequence labels $\{y_i, y_2, ..., y_n\}$
\State
\State $y \gets [o]\times n$
\For {question $q_y$ in QuestionTemplates}
  \State \{x$_{start}, ..., x_{end}$ \} = MRC(X, $q_y$)
  \State \IF $\{x_{start}, ...,  x_{end}\}\neq \textsc{NULL}$  \DO \\ 
  \hspace{0.7cm}  $\{y_{start}, ..., y_{end}\}$ = y
   \State \ENDIF
 \EndFor
 \State \Return $y$
\end{algorithmic}
\caption{Overview of the proposed model.}
\label{alg}
\end{algorithm}
\begin{table}[!ht]
\small
\center
\begin{tabular}{lll}\hline
Entity & Natural Language Question \\\hline
Facility  & Which  facility  is  mentioned  in  the  text?  \\
Location & Which location is mentioned in the text?  \\ 
Person  & Which is Person mentioned in the text?   \\\hline
\end{tabular}
\caption{Examples for transforming different entity categories to question queries. }
\label{entitytemplate}
\end{table}

\subsection{Extracting Answer Spans via MRC}
Each type of the entity is associated with a natural language question generated from templates, the details of which are shown in Table~\ref{entitytemplate}. 

Given the question $q_y$,
we need to extract text span $x_{start},...,x_{end}$ from the text $X$ given the question $q_y$ using MRC frameworks.  
We use BERT \cite{devlin2018bert} as a backbone. 
BERT utilizes large-scale pretraining based on language models and achieves SOTA results on MRC datasets like SQUAD \cite{rajpurkar2016squad}
To be in line with BERT, the question  $q_y$ and the passage $X$ are concatenated, forming the combined string  [CLS, $q_y$, SEP, X, SEP], where CLS and SEP are special tokens. 

There are  two commonly adopted strategies  for span prediction in MRC:
the first strategy \cite{seo2016bidirectional,wang2016multi} is to have two n-class classifiers to specially predict staring and ending indexes. 
The other strategy is to have n three-class classifiers: for each token $x_i\in [1,n]$,  the model predicts whether it is a start, an end or neither. 
These two strategies are the same in nature, but might lead to different  performances empirically. 
In this work, we choose the latter since  it yields better performances.\footnote{A further post-processing strategy is needed: if more than two positions within X are predicted  as starting positions, we select the one with the smallest index; if more than two positions are predicted as ending indexes, we select the one with the largest index.
The algorithm returns NULL if no starting or ending index is found. }
\subsection{Training Objective}
One of the key issue with the starting and ending index prediction for MRC tasks is the data  imbalance issue:
given the query $q_y$ and sentence $X$, there is  at most one token labeled as starting or ending, while all the rest are non-starting or non-ending. 
To deal with this issue, 
 we use dice loss  \cite{vnet} instead of cross entropy as the training objective. Dice loss is 
first
proposed for medical image segmentation tasks to handle the situation where there is a strong imbalance between the number of foreground and background pixels. It can be thought as a objective function optimizing for F score rather than accuracy, which cross entropy is approximately optimizing for. 
The dice loss can be formulated as follows:
\begin{equation}
    L_{dice}=1-\frac{2\sum_{i=0}^{n}{p_i g_i}+\lambda}{\sum_{i=0}^{n}p_i^2+\sum_{i=0}^{n}g_i^2+\lambda}
\end{equation}
wehre $p_i\in [0,1]$ denotes the starting/ending probability output from the model for $i^{th}$ token.
$g_i\in \{0,1\}$ denotes the golden probability. 
 The hyper-parameter $\lambda$ controls the trade-off between precision and recall. 
\begin{multline*}
      L_{dice}=1-\frac{2\sum_{i=0}^{L}{p_i g_i}}{\sum_{i=0}^{L}p_i^2+\sum_{i=0}^{L}g_i^2+\lambda} \\
            -\frac{\lambda}{\sum_{i=0}^{L}p_i^2+\sum_{i=0}^{L}g_i^2+\lambda}
\end{multline*}
Looking at  the  denominator of the first part, we can see that for negative examples with $g_i=0$, their $p_i$
 won't contribute. One can think this as a specific objective for recall. 
For the second part, $L_{dice}$ will still be penalized even if $g_i=0$, shooting for high precision scores.

\section{Experiments}
\subsection{Datasets and Training}
Experiments are conducted in the following datasets: 
{\bf Benchmark NER}:
We use  MSRA \cite{DBLP:conf/acl-sighan/Levow06}, Chinese OntoNotes 4.0 \cite{onto4}, and Resume dataset \cite{latticelstm2018};
{\bf Overlap NER}: We use ACE 2004 and ACE 2005 \cite{ace05}. \\
For English datasets, we use the pre-trained BERT model with cased for initialization. And text is tokenized using WordPiece Tokenizer. For Chinese datasets, we use the pre-trained BERT model. 
All hyperparameters such as learning rate, dropout and batch size are tuned using grid search on development set.

\subsection{Baseline Approaches}
We consider the following modelsm as baselines:
\begin{tightitemize}
\item {\bf Lattice LSTM:}
the word-character lattice model proposed by 
\newcite{latticelstm2018} constructs a word-character lattice. 
\item {\bf Glyce Lattice LSTM}: \newcite{wu2019glyce} utilizes glyph information of Chinese characters into Lattice LSTM model. 
\item {\bf Hyper-graph LSTM:} \newcite{kati2018nested} proposes a hypergraph-based model that uses LSTM for learning feature representations. 
\item {\bf Transition Model:} \newcite{luwei2018nest} introduces a scalable transition-based method to model the nested structure of mentions. 
\item {\bf Segmental Hypergraph Model :} \newcite{luwei2018overlap} proposes a  segmental hypergargh representation to model overlapping entity mentions. 
\end{tightitemize}

\subsection{Results and Discussions}

Table 2 presents the comparisons between our model and the current state-of-the-art NER models. For MSRA, our model outperforms fine-tuning BERT by +0.95\% in terms of F-scores, achieving the new state-of-the-art. On Chinese OntoNotes, our model achieve a huge gain of 2.95\% improvement in terms of F-score.
Resume is released by \newcite{latticelstm2018} and it contains eight fine-grained entity categories. Since queries contain semantic prior knowledge, our model enhanced the performance compared with fine-tuning BERT tagger. On ACE 2004, our model achieved state-of-the-art performance with 84.14\% in terms of F-scores. For ACE 2005, we enhance the F-score from 74.5\% to 86.88\%. 

\begin{table}
\small
\begin{tabular}{|llll|}\hline
\multicolumn{4}{|c|}{MSRA}\\\hline
Model & P & R & F \\\hline
\newcite{latticelstm2018}  & 93.57& 92.79& 93.18  \\
\newcite{wu2019glyce}& 93.86&93.92 & 93.89\\
BERT Tagger& 94.97 & 94.62 & 94.80 \\
BERT Query& {\bf 96.18} & {\bf 95.12} & {\bf 95.75} \\
&  & & {\bf (+0.95)}\\\hline\hline
\multicolumn{4}{|c|}{Resume}\\\hline
Model & P & R & F \\\hline
\newcite{latticelstm2018}  & 94.81& 94.11&  94.46  \\
\newcite{wu2019glyce}&95.72&95.63& 95.67 \\
BERT Tagger& 96.12 & 95.45 & 95.78 \\
BERT Query& {\bf 97.33} & {\bf 96.61} & {\bf 96.97} \\
&  &  & {\bf (+1.19)}\\\hline\hline
\multicolumn{4}{|c|}{Chinese OntoNotes}\\\hline
Model & P & R & F \\\hline
\newcite{latticelstm2018}  & 76.35& 71.56& 73.88  \\
\newcite{wu2019glyce}&82.06& 68.74&74.81  \\
BERT Tagger & 78.01& 80.35& 79.16 \\
BERT Query& {\bf 82.98} & {\bf 81.25} & {\bf 82.11} \\
&  &  & {\bf (+2.95)}\\\hline\hline
\multicolumn{4}{|c|}{ACE 2004}\\\hline
Model & P & R & F \\\hline
\newcite{kati2018nested} & 73.6 & 71.8 & 72.7 \\
\newcite{luwei2018nest} & 74.9 & 71.8 & 73.3 \\
\newcite{luwei2018overlap} & 78.0 & 72.4 & 75.1 \\
BERT Tagger & 79.39 & 79.97 & 79.68 \\
BERT Query& {\bf 84.05} & {\bf 84.23} & {\bf 84.14} \\
& & & {\bf (+4.46)}\\\hline\hline
\multicolumn{4}{|c|}{ACE 2005}\\\hline
Model & P & R & F \\\hline
\newcite{kati2018nested} & 70.6 & 70.4 & 70.5 \\
\newcite{luwei2018nest} & 74.5 & 71.5 & 73.0 \\
\newcite{luwei2018overlap} & 76.8 & 72.3 & 74.5 \\
BERT Tagger &78.21 & 82.74 & 80.41 \\
BERT Query& {\bf 87.16} & {\bf 86.59} & {\bf 86.88} \\
&  & & {\bf (+6.47)}\\\hline\hline
\end{tabular}
\caption{Results for NER tasks.}
\label{NER results}
\end{table}


\section{Ablation study}
\subsection{Size of Training Data}
\begin{figure}[t]
\centering
\includegraphics[scale=0.3]{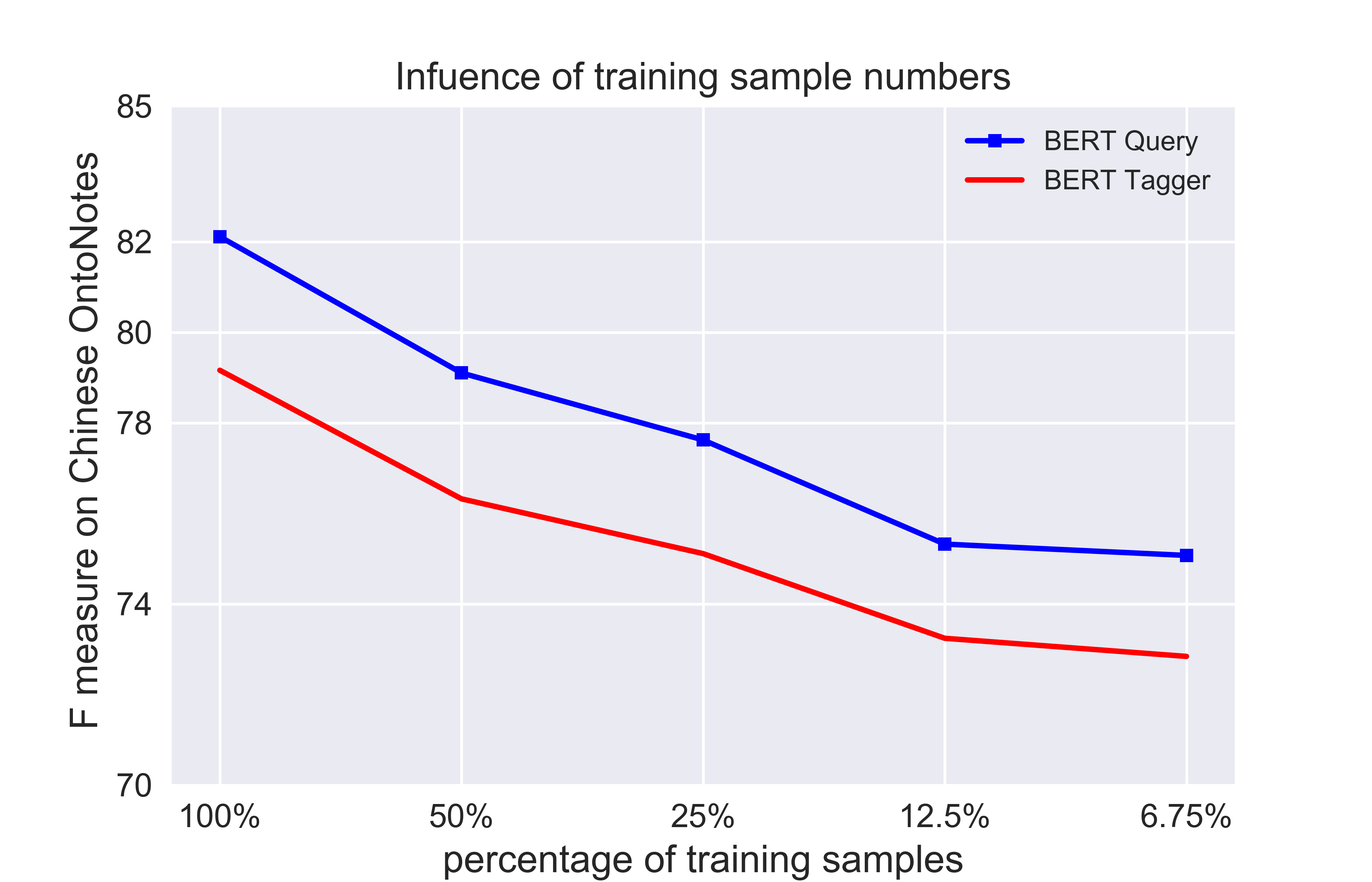}
\caption{model performance decrease with less training data}
\label{train-samples}
\end{figure}

Since the natural language query encodes significant prior knowledge, we expect that the proposed framework works better with less training data. 
Figure\ref{train-samples} verifies this point: 
on the OntoNotes training set, 
the 
query-based approach achieves comparable performance to BERT even when with half amount of training data.

\subsection{Different Query Choices}
To analyze the impact of query, we compared 3 kinds of query: 1) index query (e.g.,``one", ``two", ``three"), 2) pseudo query (e.g.,``person", ``location",``company") and 3) natural language query. 
 Performances regarding different strategies are shown in Table\ref{ablation-query}.
We find natural language queries lead to best performance due to the concrete knowledge they encode. 
\begin{table}
\small
\center
\begin{tabular}{|lll|}\hline
Datasets & loss &  f-1 score \\\hline
Chinese OntoNotes & Entropy &  80.20  \\
Chinese OntoNotes & Dice Loss & 82.11 (+1.91)\\ 
MSRA & Entropy &  95.41  \\
MSRA & Dice Loss & 95.75 (+0.34) \\ 
Resume & Entropy & 96.83    \\
Resume & Dice Loss & 96.97 (+0.14)  \\ \hline
\end{tabular}
\caption{Query Type Samples}
\label{loss-results}
\end{table}

\begin{table}
\small
\center
\begin{tabular}{|llll|}\hline
\multicolumn{4}{|c|}{Chinese OntoNotes}\\\hline
Query Type & P & R & F \\\hline
index & 81.35 & 80.92 & 81.14 \\
pseudo & 82.52 & 81.13 & 81.82 \\ 
natural & 82.98 & 81.25 & 82.11 \\\hline\hline
\end{tabular}
\caption{Results of different types of queries}
\label{ablation-query}
\end{table}

\subsection{Different Loss Functions}
We compare the performances for dice loss and cross-entropy loss in Table \ref{loss-results}.
As can be seen, dice loss yields significant performance boost than cross-entropy loss. 

\section{Conclusion}
In this paper, we reformalize the NER task as a MRC question answering task. This formalization comes with several key advantages: (1) being capable of addressing overlapping or nested entities; (2) the query encoding significant prior knowledge about the entity category to extract. This proposed strategy obtains SOTA results on five different NER datasets.

\bibliography{tagging_emnlp}
\bibliographystyle{acl_natbib}
\end{CJK*}
\end{document}